%% file: main.tex
\documentclass{article}

\usepackage{arxiv}
\usepackage{pdflscape}
\usepackage[utf8]{inputenc} 
\usepackage[T1]{fontenc}    
\usepackage{url}            
\usepackage{booktabs}       
\usepackage{amsfonts}       
\usepackage{nicefrac}       
\usepackage{makecell}
\usepackage{microtype}      
\usepackage{lipsum}
\usepackage{caption}
\usepackage{graphicx}
\usepackage{amsmath}
\usepackage{enumitem}
\graphicspath{ {./images/} }
\usepackage{float}
\usepackage{multirow}
\usepackage{tabularx}
\usepackage{algorithm}
\usepackage{algpseudocode}
\usepackage{longtable}
\usepackage{ragged2e}

\usepackage{xcolor}
\usepackage[numbers]{natbib}
\usepackage[colorlinks=true, linkcolor=blue, urlcolor=blue, citecolor=blue]{hyperref}

\title{Enhancing the KidSat Model: Integrating Geographical Encoding and Data Quality Assessment for Childhood Poverty Prediction}

\hypersetup{
    colorlinks=true,
    linkcolor=blue,
    urlcolor=blue,
    citecolor=blue
}

\author{
  Hou Hin Ip$^\star$ \\
  School of Mathematics\\
  University of Bristol\\
  \texttt{le230360@bristol.ac.uk} \\
  \And 
  Ka Nam Lam$^\star$ \\
  School of Mathematics\\
  University of Bristol\\
  \texttt{cy23765@bristol.ac.uk} \\
  \And 
  Joshua Man Yu Ng$^\star$ \\
  School of Mathematics\\
  University of Bristol\\
  \texttt{jq22530@bristol.ac.uk} \\
  \And 
  Makkunda Sharma \\
  Department of Computer Science\\
  University of Oxford\\
  \texttt{makkunda.sharma@st-hughs.ox.ac.uk} \\
  \And 
  Seth Flaxman \\
  Department of Computer Science\\
  University of Oxford\\
  \texttt{seth.flaxman@cs.ox.ac.uk} \\
  \And 
  Codie Gerlach-Wood \\
  School of Mathematics\\
  University of Bristol\\
  \texttt{codie.gerlach-wood@bristol.ac.uk} \\
  \And 
  H. Juliette T. Unwin \\
  School of Mathematics\\
  University of Bristol\\
  \texttt{juliette.unwin@bristol.ac.uk} \\
}

\renewcommand{\today}{}

\begin{document}
\maketitle

\begin{abstract}
Accurate poverty mapping using satellite imagery is often hindered by (i) noisy and sparse survey-derived supervision, (ii) image quality issues such as cloud cover and image corruption, and (iii) lack of explicit spatial structure in image-only models. Building on the KidSat framework, we develop an enhanced pipeline that improves predictive accuracy via refined data preprocessing, systematic image quality assessment, and mathematically defined geographic encoding. First, we refine the fine-tuning target matrix by resolving high-cardinality sparsity and reducing one-hot dimensionality from 103 to 51 via DHS re-aggregation. Second, we introduce a simple two-stage quality-screening procedure to filter heavily clouded or corrupted observations. Third, we fuse DINOv2 visual embeddings with Spherical Harmonics (SH) location features. Across extensive experiments, these changes reduce MAE from 0.2167 to 0.1759, corresponding to an 18.83\% relative reduction on the cluster-level severe-deprivation proportion scale. When extended from 16 to 33 African countries, the best-performing configuration achieves an overall MAE of 0.1658.
We find that SH features consistently improve performance over the image-only backbone, whereas higher-capacity coordinate Multi Layer Perception augmentation (SH+SIREN) can underperform without carefully designed objectives. Finally, gradient-boosted tree heads (XGBoost/LightGBM) most effectively exploit nonlinear interactions in the fused visual-geographic representation. These findings provide a scalable and principled recipe for improving satellite-based socioeconomic predictions using only publicly accessible data.
\end{abstract}

\section{Introduction}
\input{Sections/Introduction}

\section{Methods} 
\label{second:Methods}
\input{Sections/Methods}

\section{Experiments}
\label{third:Experiment}

\input{Sections/Experiment}

\section{Results}
\label{forth:Results}
\input{Sections/Results}

\section{Discussion}
\label{five:Discussion}
\input{Sections/Discussion}

\bibliographystyle{unsrt}
\bibliography{references}

\input{Sections/Appendix}

\end{document}

%% file: Sections/Introduction.tex
Accurate assessment of social indicators is crucial to evidence-based governance, enabling policy makers and organisations to target and evaluate public service delivery effectively \cite{HanciogluArnold2013, GrayAzzopardiKennedyWillersdorfCreati2013}. However, the data required to construct such indicators remain scarce and limited. Traditional approaches rely on large-scale household surveys, including the Demographic and Health Surveys (DHS) \cite{DHSProgram} and Multiple Indicator Cluster Surveys (MICS) \cite{MICSWebinarNatCen2024}, which are time-consuming to conduct, infrequent and in many regions depend on external funding approvals to conduct \cite{Duthe2025}. For example, between 2000 and 2016, 39 of 59 African countries conducted fewer than two nationally representative consumption surveys, demonstrating a significant data gap that prevents the timely reflection of demographic change \cite{yeh2020satellite}.

Remote sensing offers a promising complement to conventional surveys. An early and influential strand of this literature used nighttime lights as a proxy for local economic activity. Henderson et al.\ (2012)~\cite{henderson2012measuring} demonstrated that night-time satellite observations could augment official income growth measures in countries with weak national statistical systems, and Jean et al.\ (2016)~\cite{jean2016combining} showed that daytime satellite imagery, combined with convolutional neural networks (CNNs), could explain up to 75\% of asset wealth across five African countries. Subsequent work by Yeh et al.\ (2020)~\cite{yeh2020satellite} extended these methods to incorporate publicly available multispectral imagery from approximately 20,000 African villages, adequately explaining 70\% of the variation in village-level wealth in their corresponding countries. Chi et al.\ (2022)~\cite{chi2022micro} further demonstrated the global scalability of this approach by combining high-resolution satellite imagery at 2.4\,km resolution with mobile phone connectivity data and produced micro wealth estimates covering the populated surface of 135 low- and middle-income countries. Taken together, this body of work establishes that computer vision applied to freely available satellite data can provide accurate and frequently updated estimates of socioeconomic conditions~\cite{HallDompaeWahabDzanku2023}.

Building on this approach, the KidSat project \cite{Sharma2024KidSat} introduced a computer vision framework to estimate childhood poverty using 33,608 publicly available satellite images paired with DHS survey data from 16 countries in Southern and Eastern Africa. The model targets the proportion of children under the age of 18 within a region who experience severe deprivation, defined by UNICEF as deprivation in at least one of six dimensions: health, nutrition, water, education, sanitation, or housing \cite{UNICEF2019}. To construct this target variable, 17 key DHS variables are combined to generate a binary indicator for each child, denoting whether they fulfill the criteria for severe deprivation or not. Individuals residing within the same geographic area are then aggregated into sampling units known as clusters \cite{DHSProgram2025}, expressed on a continuous $[0,1]$ scale, serves as the prediction target.

To learn predictive representations from satellite imagery, KidSat employs self-supervised computer vision architectures. The original study evaluated several models, including DINOv2~\cite{DINOv22023}, SatMAE~\cite{SatMAE2022}, and MOSAIKS~\cite{rolf2021mosaiks}, finding that the DINOv2 Vision Transformer (ViT) achieved the lowest mean absolute error (MAE) across both spatial and temporal benchmarks. Under this framework, each satellite image was processed through the DINOv2 encoder and fine-tuned against the 17 DHS variables to output a 768-dimensional feature vector, which was subsequently passed through a linear regression head to estimate the cluster-level proportion of severe deprivation.

Despite these advances, we identify three limitations in the original pipeline that constrain predictive accuracy. First, the fine-tuning matrix suffers from high sparsity, arising from the one-hot encoding of high-cardinality categorical DHS variables, which weakens gradient signals and introduces numerical instability. Second, the original pipeline lacks a systematic image quality selection procedure, defaulting to the first available satellite image per cluster regardless of cloud contamination or sensor corruption, which introduces noise and causes replication inconsistencies across computational environments. Third, the model exhibits a robustness limitation due to its exclusive reliance on visual features. The lack of integration of complementary spatial information constrains its ability to capture location-specific patterns and dependencies, thereby reducing resilience to variability in visual inputs and limiting generalisation across heterogeneous regions.

To address these limitations, we propose an enhanced data-processing pipeline that improves predictive accuracy through three targeted contributions. (1)~We refine the fine-tuning target matrix by resolving high-cardinality sparsity via informed DHS re-aggregation, incorporating two additional socioeconomic predictors (the rural--urban residence indicator and the household wealth index), and reducing one-hot dimensionality from 103 to 51 columns. (2)~We introduce a two-stage image quality screening procedure, combining scan-line gap detection with physics-based cloud detection inspired by FMASK \cite{ZHU2012, ZHU2015}, to systematically identify and replace corrupted or heavily cloud-affected observations. (3)~We augment the DINOv2 visual embeddings with Spherical Harmonics (SH) geographic encodings, providing structured spatial context that complements image-based features. 

This paper is structured as follows. Section~\ref{second:Methods} details our methodological improvements. Section~\ref{third:Experiment} presents the experimental setup, and Sections~\ref{forth:Results} and~\ref{five:Discussion} report experimental results and discuss their implications.

%% file: Sections/Methods.tex
We propose two complementary methodological enhancements to improve the robustness of the KidSat framework. First, we refine the original dataset and preprocessing pipeline to address limitations in feature representation, sparsity and satellite image quality. Second, we incorporate geographical location information through SH and SIREN encoder to provide additional spatial context that is not captured by satellite imagery alone. These improvements aim to reduce the model’s sensitivity to data quality issues and enhance its predictive performance.

\subsection{Refinement of the Fine-tuning Matrix}
The original KidSat paper fine-tuned the model solely on the 17 DHS variables that construct the severe deprivation target. Using one-hot encoding (OHE) \cite{Rodriguez2018}, this expanded into a 103-dimensional matrix with high sparsity, where 50 of the initial 103 dimensions contained fewer than 5\% non-zero entries, weakening gradient signals \cite{CerdaVaroquauxKegl2018, CerdaVaroquaux2019} and causing numerical instability during normalisation \cite{CerdaVaroquaux2019}. We address this in two ways. First, we augment the variable set with the rural--urban residence indicator \cite{Poirier2020} and the household wealth index \cite{PokhriyalJacques2017}, both of which exhibit strong correlations with poverty in prior work \cite{LI2022}. Second, rather than applying generic frequency-based grouping \cite{Potdar2017}, we implied an informed re-aggregation strategy that manually merging sparse categorical columns based on evidence from DHS codebooks \cite{DHSProgram_Recode7_2018} and prior literature \cite{Trussell2011, Demoze2025}; where no empirical justification existed, rare categories were consolidated into a single ``Other'' class. These changes reduce the fine-tuning matrix from 103 to 48 dimensions. The full details of the retained dimensions are provided in Appendix~\ref{AppendixB}.

\subsection{Image Quality Assessment and Selection}
\begin{figure}[H]
    \centering
    \includegraphics[width=0.8\linewidth]{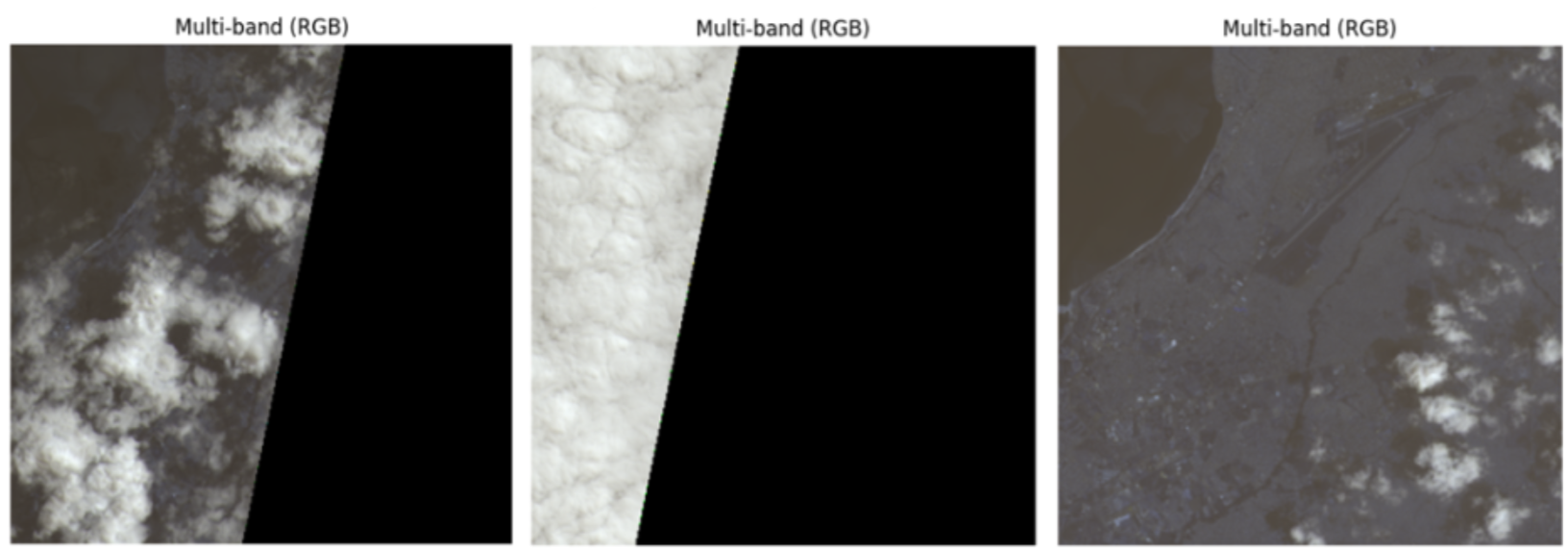}
    \caption{Three multi-band (RGB) images from Luanda, Angola (early 2015) captured at 16–18 day intervals. The left and middle images are corrupted by heavy cloud cover and dark pixels, limiting their predictive information.}
    \label{fig:corrupted-images}
\end{figure}

While the original KidSat paper utilised both Sentinel-2 and Landsat 5/7/8 imagery, our study focuses solely on Landsat 7/8 imagery with images at a resolution of $336 \times 336$ pixels. The original pipeline defaulted to select the first available image per cluster regardless of degradation, which caused replication inconsistencies across GPU configurations (MAE variation $< 0.0005$). Figure~\ref{fig:corrupted-images} illustrates the two principal degradation sources affecting Landsat 7/8 imagery: null pixels arising from scan-line errors \cite{Hossain2015} and cloud cover contamination.

We developed a two-stage detection algorithm to address these issues. Stage one identified sensor corruption by detecting scan-line gaps in Landsat 7 imagery and null pixels (RGB $= (0,0,0)$) in Landsat 7/8 imagery. Stage two implemented physics-based cloud detection, inspired by FMASK \cite{ZHU2012, ZHU2015}, to compute four spectral indices to characterise cloud presence: brightness, NDVI, blue-to-red ratio, and whiteness index. Adaptive percentile-based thresholds were then applied, set at the 85th percentile for Landsat 7 and the 90th percentile for Landsat 8 to reduce false positives over bright urban areas and to reflect the higher radiometric sensitivity of Landsat 8. Morphological filtering was applied to refine the resulting binary cloud masks \cite{QIU2019}.

We computed an image quality metric based on the proportion of degraded pixels that integrated both stages. Applying a 30\% degradation threshold, approximately 15\% of images were replaced or excluded, with the lowest-degradation image selected per cluster to ensure consistency across training and evaluation phases.

\subsection{Integrating Geographical Encoding}
Satellite imagery alone may be insufficient to capture broader spatial and socioeconomic context, particularly under conditions of degraded image quality. Early studies demonstrated that high-resolution daytime imagery, when coupled with appropriate supervision signals, enables convolutional representations to capture meaningful socioeconomic gradients \cite{jean2016combining}. Subsequent work using publicly available multispectral imagery further emphasised the role of spatial structure in enhancing village-level predictions and improving cross-country generalisation \cite{yeh2020satellite}. Motivated by these findings, we incorporate explicit location information through a dedicated encoding mechanism to complement visual features and enhance model robustness.

Representing location is non-trivial at a global scale because geographic coordinates lie on a spherical surface rather than a Euclidean plane. A simple choice is to use raw latitude and longitude, but this ignores essential spherical geometry and periodicity. Prior work has explored a range of encoding strategies to address these limitations. Fixed sinusoidal positional encodings popularised in Transformers \cite{vaswani2017attention} and related Fourier feature mappings \cite{tancik2020fourier} inject a multi-scale periodic structure, which aids models to recover high-frequency signals. Random Fourier Features (RFF) approximate stationary kernels with finite-dimensional embeddings suitable for linear models \cite{rahimi2007random}. Learning-based approaches further extend these ideas: Space2Vec learns multi-scale periodic encodings inspired by grid cells \cite{mai2020space2vec}, while Sphere2Vec operates directly on the $S^2$ manifold to preserve great-circle distances, thereby improving robustness in polar and data-sparse regions \cite{mai2023sphere2vec}. Building on this landscape of methods, and following recent evaluations by  Rußwurm et al. \cite{russwurm2023geoencoding}, we adopt two geographic encoders: Spherical Harmonics (SH) and a hybrid SH+SIREN model.



\paragraph{SH encoding.}
Given geographic coordinates $(\theta, \phi)$, the SH encoder computes the real and imaginary parts of spherical harmonics $Y_\ell^m$ up to degree~$L=15$, producing a fixed-dimensional feature vector that captures multi-scale spatial structure without any learnable parameters.

\paragraph{SH+SIREN encoding.}
The hybrid encoder passes SH features through a SIREN network, which defaults to four hidden layers with 256 neurons each and sinusoidal activations, producing a 128-dimensional learned geographic embedding. 
Training follows two stages:


\begin{enumerate}
    \item Pre-training Stage: Given the geographic coordinates, the SH encoder generates a vector that contains harmonic with different degrees. The SIREN network is pre-trained to predict poverty indicators using the harmonic vector to learn location-specific spatial patterns and relationships. 
    \begin{equation}
    \text{geo} \;=\; \operatorname{SIREN}\!\big(\,[\Re Y_{\ell}^{m},\Im Y_{\ell}^{m}]_{\ell\le L}\big).
    \end{equation}
    \item Feature Fusion Stage: The learned geographic embeddings from the pre-trained SH+SIREN network are concatenated with visual features extracted from satellite imagery:
    \begin{equation}
    z \;=\; \bigl[\, \text{geo(loc)} \;\| \; \text{(image embedding)} \,\bigr].
    \end{equation}
\end{enumerate}

\begin{figure}[t]
    \centering
    \includegraphics[width=0.8\textwidth]{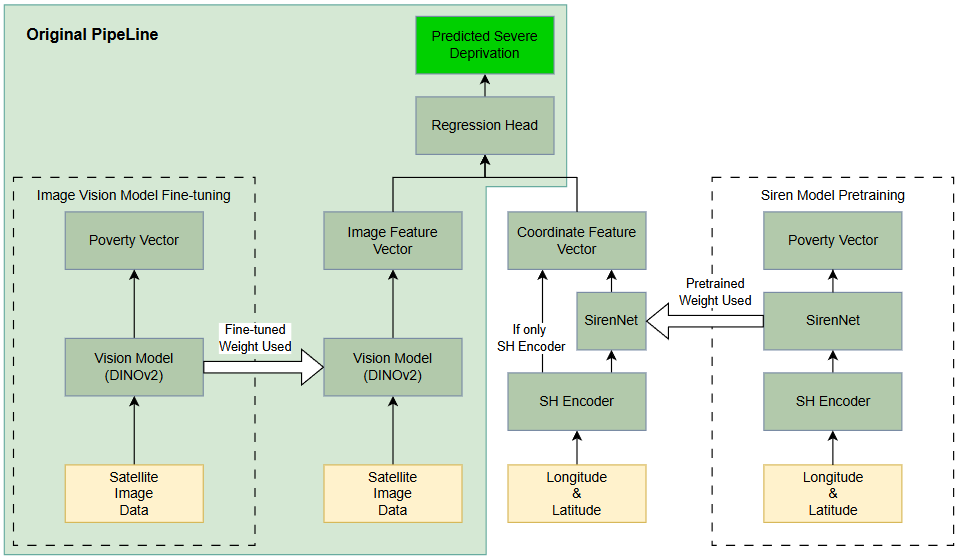}
    \caption{The Extended KidSat Model Architecture. Left: DINOv2 is fine-tuned against the poverty vector. Right: the SH+SIREN network is pre-trained on poverty indicators. Centre: the resulting visual and geographic embeddings are concatenated and passed through a regression head to predict severe deprivation.}
    \label{fig:kidsat-arch}
\end{figure}

The fused representation $z$ is then passed to a regression head for severe deprivation estimation as shown in Figure~\ref{fig:kidsat-arch}. When using SH encoding alone, the raw SH feature vector is concatenated directly with the visual embedding, bypassing the pre-training stage.

%% file: Sections/Experiment.tex
\subsection{Experimental Setup}

We evaluate the effect of geographic encoding through an ablation study comparing three configurations: the image-only DINOv2 baseline, DINOv2 with SH features, and DINOv2 with the hybrid SH+SIREN encoder. This setup allows us to isolate whether explicit geographic information improves prediction performance beyond visual features alone.

We follow the 5-fold cross-validation protocol used in KidSat: 5-fold cross-validation is applied across the entire dataset. For every fold, DINOv2 fine-tuning, geographic encoder training and the regression head are trained only on 80\% of the clusters, while the remaining 20\% held-out fold is reserved exclusively for evaluation. The same fold partitions are used across all three experiments to ensure a fair comparison. All experiments are run using an NVIDIA A40 GPU, and performance is assessed using mean absolute error (MAE) for fine-tuning and evaluation. Since the prediction target is the cluster-level proportion of severely deprived children, all MAE values are reported on a $[0,1]$ prevalence scale.
Further hyperparameter configurations can be found at Appendix \ref{AppendixD}.

\subsection{Geographic Encoder Ablation}


We conduct three experiments to isolate the effect of each geographic encoding component.
These ablation experiments are performed on the original KidSat country set, before extending the best-performing configuration to the expanded 33-country dataset.

\begin{enumerate}
    \item Baseline (DINOv2 only): The original KidSat pipeline based on DINOv2 only.
    \item SH only: DINOv2 visual features concatenated with spherical harmonics geographic embeddings.
    \item SH + SIREN: DINOv2 visual features concatenated with SH + SIRENNET embeddings.
\end{enumerate}

For each experiment, the fused feature representation is passed to a cross-validated ridge regression model with a sigmoid link function to estimate the probability of severe deprivation. The details of how SH and SIREN embeddings are generated are provided in the two-stage training subsection and are not repeated here.

Each of the three experiments follows the same overall process:
\begin{enumerate}
    \item Fine-tune DINOv2 using the DHS poverty factor.
    \item Generate geographic embeddings according to the experiment configuration.
    \item Concatenate features to form the fused representation.
    \item Train a cross-validated ridge regression model.
    \item Evaluate MAE on held-out folds.
\end{enumerate}

\subsection{Regression Heads Comparison}
While the original KidSat pipeline used a simple linear regression head (e.g., ridge regression), the fused space introduces non-trivial cross-modal interactions that linear models cannot capture without explicitly engineering interaction terms. In our setting, the same visual cue can carry different poverty-related meanings across geographic contexts. For example, "sparse vegetation" (captured by DINOv2) is typical in arid regions (e.g., the Sahel) and therefore weakly informative about poverty level, whereas in humid regions (e.g., the Congo Basin) unusually sparse vegetation can indicate soil degradation or limited irrigation and thus correlate with higher deprivation. These conditional relationships are inherently multiplicative or non-linear, motivating predictors that can learn interactions directly.

Beyond the ridge-regression baseline used in prior work, and motivated by the limitation of a linear head, we assessed a broader family of predictors on the fused vector:
\begin{itemize}
  \item \textbf{Ridge regression (baseline):} A linear predictor with $\ell_2$ regularization, providing a well-regularized and interpretable point of comparison and a lower bound for performance without cross-modal interactions.
  \item \textbf{Tree ensembles (Random Forest \cite{breiman2001randomforests}, LightGBM \cite{ke2017lightgbm}, XGBoost \cite{chen2016xgboost}):} These models naturally capture piecewise non-linear relationships and feature interactions without manual engineering. In the fused space, they can reweight visual cues conditionally on geography (e.g., downweighting sparse vegetation as a poverty indicator in arid zones where it is geographically normal). Their bagging/boosting regularization also suits the high dimensionality of $z$.
  \item \textbf{Shallow Deep Neural Network (3/4 layer multi layer perception (MLP)):} A small MLP provides smooth, parametric modelling of non-linear interactions with small capacity and standard regularization.
\end{itemize}

In this set of experiments, different regression heads are evaluated by replacing the ridge regression model used in the geographic encoder experiments. The fused embedding generated from the selected encoder configuration remains fixed, and only the final regression component is varied. Each regression head is trained using five-fold cross-validation, and MAE is computed on the held-out validation folds to ensure a fair and consistent comparison.

\subsection{Extension to additional countries in Africa}
We expanded the original KidSat dataset from 16 countries (46 surveys) in Southern and Eastern Africa to 33 countries total, adding 16 countries (33 surveys) from Central and Western Africa, yielding 43,823 clusters for prediction. The best-performing configuration (SH + LightGBM with improved preprocessing) was applied to this expanded dataset.

%% file: Sections/Results.tex
\subsection{Results for geographic encoder ablation and regression heads}

Table~\ref{tab:encoder-results} summarises model performance across the three geographical encoder configurations and five regression head settings. Our model replicates the original KidSat training pipeline as the baseline for comparison, achieving an MAE of $0.2167~(\pm~0.0013)$. Refining the fine-tuning matrix through variable augmentation and re-categorisation of sparse categorical features, combined with enhanced image quality assessment, reduced the MAE to $0.1980~(\pm~0.0008)$. This corresponds to an absolute reduction of $0.0187$ on the severe-deprivation proportion scale, or an $8.63\%$ relative reduction compared with the baseline.

At the same time, incorporating geographic information further improved prediction accuracy. Under the configuration without processing improvement, adding the SH encoder alone produced the strongest gains across all regression heads, achieving MAEs between $0.1833~(\pm~0.0007)$ (XGBoost) and $0.2031~(\pm~0.0010)$ (Ridge), with XGBoost yielding the largest absolute improvement of $0.0334$ relative to the baseline. By contrast, the SH + SIREN encoder yielded only marginal improvements, with MAEs ranging from $0.2110~(\pm~0.0009)$ (XGBoost) to $0.2149~(\pm~0.0011)$ (MLP). These trends also held if data and imagery processing improvement is performed with SH encoding again led to the largest performance gains, with the best MAE of 0.1759 ± 0.0016 (LightGBM), representing a 0.0221 reduction compared to the improved baseline. This confirm that harmonic features provide consistent predictive benefit across model classes.

Surprisingly, the SH + SIREN encoder underperformed plain SH in nearly all settings. Under improved imagery and preprocessing, the best SH + SIREN result reached an MAE of 0.1853 ± 0.0022 using XGBoost, which is 0.0094 higher than the corresponding SH-only configuration. In contrast, SH encoding alone consistently achieved the strongest performance across regression heads. LightGBM combined with SH yielded the lowest observed MAE of 0.1759 ± 0.0016. Tree-based models, including XGBoost and LightGBM, achieved lower errors than both linear and neural regression heads. Ridge regression, used as the standardised evaluation model, produced weaker results across all encoder settings.

Across regression heads, the tree-based models (XGBoost and LightGBM) consistently achieved the best performance. When combined with SH encoding, LightGBM yielded the lowest observed MAE of 0.1759 ± 0.0016, highlighting the advantage of modeling nonlinear interactions between visual and geographic features. Random Forest also outperformed linear models but lagged behind gradient-boosted approaches. Ridge regression, used as the standardised evaluation head, showed stable but weaker performance, suggesting that geographic context introduces nonlinear feature interactions that linear models cannot fully capture.

\input{Tables/geo_encoding_results}

 \begin{figure}[t]
    \centering
    \includegraphics[width=0.7\textwidth]{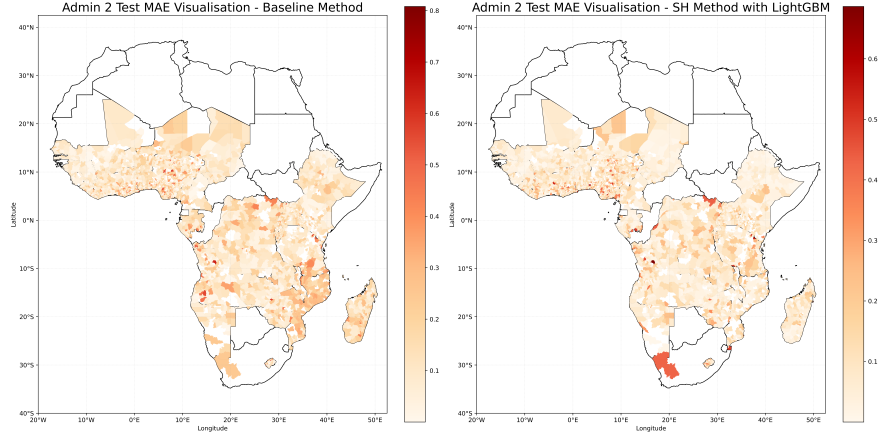}
    \captionsetup{justification=raggedright, singlelinecheck=false}
    \caption{Visualisation demonstrating comparison between Baseline method (Left) and visualisation with modified process (Right), with a total of 33 African countries.}
    \label{fig:comparison-maps}
\end{figure}

\subsection{Results for extending to additional countries in Africa}

By combining the improved data pre-processing strategy with spherical harmonic encoding and the LightGBM model, the training was expanded to include 17 additional countries and 33 surveys comprising 20,271 new clusters. This configuration achieved an MAE of 0.1658 ($\pm$ 0.0005), demonstrating the robustness and consistency of the model across geographically diverse regions (Figure~\ref{fig:comparison-maps}).

%% file: Tables/geo_encoding_results.tex
\begin{table}[t]
    \caption{Comparison of MAEs on regression heads under different geo-encoder and regression head settings.}
    \centering
    \renewcommand{\arraystretch}{1.5}
    \begin{tabular}{|l|l|c|c|}
        \hline
        \multicolumn{2}{|c|}{\makecell{\textbf{Encoder / Regression Head} \\ \textbf{evaluated on A40 GPU}}} & 
        \makecell{\textbf{Without Improvement on} \\ \textbf{Data and Imagery Processing}} &
        \makecell{\textbf{With Improvement on} \\ \textbf{Data and Imagery Processing}} \\
        \hline
        \multicolumn{2}{|l|}{Without Geo Encoder} & 
        0.2167 $\pm$  0.0013 & 0.1980 $\pm$  0.0008 \\
        \hline
        \multirow{5}{*}{\textbf{Adding SH + SIREN}} & Ridge & 0.2154 $\pm$  0.0008 & 0.1885 $\pm$  0.0018 \\
         & XGBoost & 0.2110 $\pm$  0.0009 & 0.1853 $\pm$  0.0022 \\
         & LightGBM & 0.2110 $\pm$  0.0014 & 0.1847 $\pm$  0.0018 \\
         & Random Forest & 0.2146 $\pm$  0.0010 & 0.1855 $\pm$  0.0018 \\
         & MLP & 0.2149 $\pm$  0.0011 & 0.1897 $\pm$  0.0013 \\
        \hline
        \multirow{5}{*}{\textbf{Adding SH Encoder}} & Ridge & 0.2031 $\pm$  0.0010 & 0.1888 $\pm$  0.0011 \\
         & XGBoost & 0.1833 $\pm$  0.0007 & 0.1771 $\pm$  0.0020 \\
         & LightGBM & 0.1835 $\pm$  0.0006 & \textbf{0.1759 $\pm$  0.0016} \\
         & Random Forest & 0.1888 $\pm$  0.0010 & 0.1797 $\pm$  0.0018 \\
         & MLP & 0.1902 $\pm$  0.0007 & 0.1823 $\pm$  0.0024 \\
        \hline
    \end{tabular}
    \label{tab:encoder-results}
\end{table}

%% file: Sections/Discussion.tex
Building on the KidSat project's demonstration that combining satellite imagery with household survey data can predict socioeconomic outcomes, our study establishes an enhanced methodology that improves predictive accuracy through refined data preprocessing, systematic image quality assessment, and explicit geographic encoding. Specifically, by reducing fine-tuning dimensionality and incorporating explicit geographic context via SH, our framework achieved a relative 18.83\% reduction in MAE compared with the baseline model. The approach also maintained robust performance when scaled from 16 to 33 African countries, achieving an overall MAE of 0.1658.

\subsection{Data Preprocessing and Image Selection}

Our key findings indicated that the quality of the fine-tuning matrix remained fundamental constraints on model performance. While UNICEF’s six deprivation dimensions underpinned the predictors, optimal performance required incorporating correlated but distinct DHS variables, such as the rural–urban indicator and household wealth index. Therefore, the selection of fine-tuning variables needed to be guided by independent empirical justification rather than limited to components defining the target outcome. 


Furthermore, our re-aggregation strategy, informed by DHS codebooks and prior literature, successfully improved MAE without information loss. To further strengthen the fine-tuning matrix, future work could employ machine learning algorithms such as GLM and XGBoost to identify the most informative DHS predictors \cite{QIU2019}, and evaluate alternative mathematical encoding strategies, such as adoption of Gamma-Poisson matrix factorization and in-hash encoding
\cite{CerdaVaroquaux2019} to mitigate sparsity.

Finally, we illustrated that image quality assessment was essential for ensuring data reliability. When 15\% of available images exhibited over 30\% cloud or corruption coverage, poor image selection reduced accuracy by introducing noise or meaningless information. Further research could investigate restoration techniques, such as image inpainting based on deep image prior \cite{PetrovskaiaJanaOseledets2022, CzerkawskiUpadhyayDavison2022, UlyanovVedaldiLempitsky2017}, or diffusion prior \cite{ho2020denoisingdiffusionprobabilisticmodels} to enhance performance by recovering partially degraded imagery.

\subsection{Geo-encoding and Feature Fusion}

Integrating publicly accessible information, such as geographic coordinates, during model training significantly enhanced predictive accuracy by providing supplementary contextual information. This multi-source feature integration framework enhanced coverage in under-surveyed regions and mitigated data gaps arising from incomplete surveys or limited information due to low-resolution satellite imagery. 
Across all experiments for geo-encoder, the spherical harmonics (SH) encoder consistently improved performance over the image-only DINOv2 baseline, indicating that structured, low-frequency geographic priors provide complementary information to the visual backbone. In contrast, the combined SH+SIREN encoder underperformed relative to SH alone. This behaviour suggests that the SIREN network, when trained using the same poverty vector that supervises DINOv2 fine-tuning, may have introduced representation overlap rather than supplying an independent spatial signal, thereby reducing the overall complementarity of the fused representation.

The consistently strong performance of tree-based regression heads, particularly XGBoost and LightGBM, indicates that the fused visual–geographic representation contains meaningful nonlinear interactions that linear models cannot fully exploit. This accords with prior evidence that gradient boosting methods are well-suited to heterogeneous, high-dimensional feature fusion settings. Compared with neural regressors, boosted decision trees have been shown to capture nonlinear interactions effectively while maintaining robustness in limited-data regimes \cite{entropy2021hyperspectral}. This likely explains their superior performance when integrating DINOv2 visual embeddings with SH geographic features.

Looking forward, multiple avenues could address the limitations observed. One direction is to revisit the SIREN training objective: training it on auxiliary demographic variables rather than the same poverty vector used by the vision model may yield more complementary spatial representations. Future research should systematically explore additional low-cost and accessible datasets that correlate with the underlying causes of poverty, including nighttime satellite imagery \cite{DollMullerMorley2006}, conflict event repositories (e.g., the Uppsala Conflict Data Program) \cite{UCDP2025}, to further enrich this integrated prediction framework.

Another direction is to explore alternative geospatial representation learning methods. Contrastive coordinate–image alignment approaches such as GeoCLIP \cite{geoclip2023} provide continuous geo-embeddings that could be fused with DINOv2 features. Retrieval-augmented neural fields such as RANGE \cite{range2025} supply multi-resolution geospatial priors that may complement both low- and high-frequency encoders. Meanwhile, LocDiff \cite{locdiff2025} demonstrates the benefit of constructing coordinate embeddings in structured Hilbert spaces to avoid instability and representational collapse. Exploring these geo-encoder families may yield richer spatial representations and reduce redundancy when combined with vision backbones.

Overall, the results indicate that well-structured geographic priors such as spherical harmonics can meaningfully improve satellite-based poverty prediction, whereas high-capacity coordinate MLPs require carefully designed objectives to avoid duplicating visual information and overfitting. The strong performance of tree-based regression heads further highlights the value of robust, nonparametric models for integrating visual and spatial modalities. Future work incorporating modern geo-encoding methods, multi-scale spatial priors, and alternative training objectives may produce even more expressive and stable poverty-prediction systems.

%% file: Sections/Appendix.tex
\newpage
\appendix
\renewcommand{\thetable}{\Alph{table}}
\setcounter{table}{0}
\section*{Appendix}

\section{DHS Variables Used in the KidSat Setup}
\label{AppendixA}
The table below presents the 17 DHS variables used in the original setup, as well as the new variables introduced in the modified configuration:

\input{Tables/DHS_variables}

\begin{landscape}
    \section{Re-aggregation of One-Hot Encoding Columns}
    \label{AppendixB}
    
    Note: In DHS surveys, each indicator corresponds to the response to a particular question, and each response is encoded using the unique DHS codebook\cite{DHSProgram_Recode7_2018}, with interpretations that are standardised across all participating countries.
    
    \begin{table}[H]
        \scriptsize
        \centering
        \renewcommand{\arraystretch}{1.25}
        \setlength{\tabcolsep}{3pt}
        \setlist[itemize]{nosep, left=8pt} 
        
        \begin{tabular}{|m{2.2cm}|m{2.2cm}|m{1.5cm}|m{1.5cm}|m{1.5cm}|p{13cm}|}
            \hline
            \textbf{Variables} &
            \textbf{Type} &
            \begin{tabular}[c]{@{}c@{}}\textbf{Sparse}\\\textbf{Columns}\\($\varepsilon$ = 0.05)\end{tabular} &
            \begin{tabular}[c]{@{}c@{}}\textbf{No. of}\\\textbf{Columns}\end{tabular} &
            \begin{tabular}[c]{@{}c@{}}\textbf{Final}\\\textbf{Columns}\end{tabular} &
            \textbf{Adjustment} \\
            \hline
            
            hv201 & Categorical & 16 & 26 & 5 &
            Adjustment of the main drinking water source, in which the following categories are merged:
            \begin{itemize}
                \item \texttt{hv201\_protected\_pipes}: [11, 12, 13, 14]
                \item \texttt{hv201\_protected\_wells}: [21, 22, 23, 24, 25, 31]
                \item \texttt{hv201\_protected\_spring}: [41]
                \item \texttt{hv201\_protected\_others}: [51, 61, 62, 63, 71]
                \item \texttt{hv201\_unprotected\_source}: [32, 33, 34, 35, 36, 42, 43, 44, 45, 46]
            \end{itemize} \\ 
            \hline
            
            hv205 & Categorical & 17 & 22 & 5 &
            Adjustment of the variable \texttt{hv205}, which is the type of toilet facility based on whether it is a protected facility:
            \begin{itemize}
                \item \texttt{hv205\_protected\_flush}: [11–19]
                \item \texttt{hv205\_protected\_ventilated\_pit\_latrine}: [21]
                \item \texttt{hv205\_protected\_pit\_latrine}: [22, 24–29]
                \item \texttt{hv205\_protected\_composting\_toilet}: [41]
                \item \texttt{hv205\_unprotected\_source}: [23, 31, 42, 43, 96]
            \end{itemize} \\ 
            \hline
            v312 & Categorical & 14 & 15 & 4 &
            Adjustment of the variable \texttt{v312}: current contraceptive method.  
            Defined groups of contraceptive use are created as follows:
            \begin{itemize}
                \item \texttt{v312\_no\_contracept}: [0] (no method used)
                \item \texttt{v312\_trad\_contracept}: [8, 9, 10] (traditional methods)
                \item \texttt{v312\_ineffective\_contracept}: [15, 16, 18] (ineffective methods)
                \item \texttt{v312\_contracept}: [1–7, 11–14, 17, 19, 20] (modern or other methods)
            \end{itemize} \\ 
            \hline
            h3& Categorical & 0 & 4 & 3 &
            {\multirow{5}{13cm}{\text{Variables} \texttt{h3}, \texttt{h5}, \texttt{h7}, and \texttt{h9} represent vaccination status. 
            In the original data pipeline, these statuses were one-hot encoded differently, 
            even though the questionnaire options are identical. 
            To maintain consistency, vaccination status is refined as follows:
                \begin{itemize}
                    \item \text{[0, 8]} — Not vaccinated / Don’t know
                    \item \text{[1, 3]} — Vaccinated (confirmed)
                    \item \text{[2]} — Reported by mother
                \end{itemize}
            \vspace{25pt}   
            } }\\ 
            \cline{1-5}
            h5& Categorical& 0 & 4 & 3 & \\ \cline{1-5}
            h7& Categorical& 0 & 4 & 3 & \\ \cline{1-5}
            h9& Categorical& 2 & 6 & 3 & \\ 
             & &  & & & \\ 
            \hline
            
            hv106 & Categorical & 0 & 1 & 4 & Categorical column — no need for re-aggregation. \\ 
            \hline
            h10 & Categorical & 0 & 2 & 2 & Categorical column — no need for re-aggregation. \\ 
            \hline
            h31 & Categorical & 0 & 2 & 2 & Categorical column — no need for re-aggregation. \\ 
            \hline
            hv109 & Categorical & 0 & 6 & 6 & Categorical column — no need for re-aggregation. \\ 
            \hline
            hv121 & Categorical & 0 & 3 & 3 & Categorical column — no need for re-aggregation. \\ 
            \hline
            hv025 & Categorical & 0 & 2 & 2 & Categorical column — no need for re-aggregation. \\ 
            \hline
            
            hv216 & Numerical & 0 & 1 & 1 & Numerical column — no one-hot encoding applied. \\ 
            \hline
            hv204 & Numerical & 0 & 1 & 1 & Numerical column — no one-hot encoding applied. \\ 
            \hline
            hv225 & Numerical & 0 & 1 & 1 & Numerical column — no one-hot encoding applied. \\ 
            \hline
            hc70 & Numerical & 0 & 1 & 1 & Numerical column — no one-hot encoding applied. \\ 
            \hline
            hv271 & Numerical & 0 & 1 & 1 & Numerical column — no one-hot encoding applied. \\ 
            \hline
            hv270 & Numerical & 0 & 1 & 1 & Numerical column — no one-hot encoding applied. \\ 
            \hline
        
        \end{tabular}
         \caption{Summary of DHS Variables Used for Fine-Tuning and Adjustments}
    \end{table}
\end{landscape}

\section{List of Countries Included in the KidSat Model}
\label{AppendixC}
\input{Tables/kidsat_country}

\section{Hyperparameters and Settings for Geographical Encoding  Evaluation}
\label{AppendixD}

This section summarises the hyperparameters and settings used for the visual backbone, geographic encoders and downstream regression heads. All experiments were performed on an NVIDIA A40 GPU to maintain consistent training behaviour across cross-validation runs with \texttt{seed = 42}  used to ensure reproducibility. All configuration and training scripts are available in our public \href{https://github.com/Picaazi/KidSat}{GitHub} repository for reference and replication.

\subsubsection{DINOv2 Fine-tuning}

We fine-tune the \texttt{dinov2\_vitb14} backbone, which outputs a 768-dimensional embedding, together with a linear prediction head with sigmoid activation. The backbone and head are optimised jointly within each cross-validation fold using the following settings:

\begin{itemize}
    \item learning rate: $1\times 10^{-6}$
    \item weight decay: $1\times 10^{-6}$
    \item optimizer: Adam (PyTorch defaults for \texttt{betas} and \texttt{eps})
    \item loss: MAE (L1 loss)
    \item epochs: $20$
    \item batch size: $8$
    \item reproducibility: \texttt{seed = 42}
\end{itemize}

No learning-rate scheduling or additional regularisation is applied.

\subsubsection{SH Encoder}
The SH encoder computes the real and imaginary parts of spherical harmonics up to degree $L=15$, producing a fixed $(L+1)^2\times2$-dimensional feature vector. This encoder has no learnable parameter.

\subsubsection{SIREN-based Spatial Encoder}

The SIREN encoder receives spherical harmonics (SH) features and produces a 128-dimensional learned spatial embedding. The encoder uses four hidden layers of width 256 with sinusoidal activation $\sin(\omega_0 x)$ where $\omega_0 = 30$.

Weights follow standard SIREN initialisation:
\begin{itemize}
    \item first layer: weights are initialized uniformly in $\left[-\frac1{n_\text{in}}, \frac1{n_\text{in}}\right]$ 
    \item hidden layers: weights are initialized uniformly in$\left[-\frac{\sqrt{6/n_\text{in}}}{\omega_0}, \frac{\sqrt{6/n_\text{in}}}{\omega_0 }\right]$
\end{itemize}

The SIREN and its prediction head are trained jointly using:

\begin{itemize}
    \item learning rate: $1\times 10^{-4}$
    \item weight decay: $1\times 10^{-6}$
    \item optimizer: Adam
    \item loss: L1 loss
    \item epochs: $200$
    \item batch size: $32$
    \item scheduler: \texttt{ReduceLROnPlateau} (factor $0.5$, patience $10$, monitored on validation MAE)
    \item gradient clipping: $\texttt{max\_norm} = 1.0$
    \item early stopping: triggered when LR $< 1\times 10^{-7}$
\end{itemize}

The resulting 128-dimensional embedding is concatenated with DINOv2 features for downstream regression.

\subsubsection{Regression Heads on Learned Representations}

For experiments on extracted representations, we evaluate a range of regression models. All models are trained on \texttt{StandardScaler}-normalised features and evaluated using 5-fold cross-validated MAE.

\paragraph{Ridge Regression}

\begin{itemize}
    \item implemented using \texttt{RidgeCV}
    \item alphas: $\{10^{-6}, \ldots, 10^{6}\}$ (log-spaced)
    \item cross-validation: 5-fold
    \item scoring: negative MAE
\end{itemize}

\paragraph{Tree-based Models}

\begin{itemize}
    \item \textbf{XGBoost}: \texttt{XGBRegressor}
          \begin{itemize}
              \item $n_{\text{estimators}} = 1000$, $max\_depth = 6$
              \item $learning\_rate = 0.1$
              \item $subsample = 0.8$, $colsample\_bytree = 0.8$
              \item $random\_state = 42$
          \end{itemize}

    \item \textbf{LightGBM}: \texttt{LGBMRegressor}
          \begin{itemize}
              \item $n_{\text{estimators}} = 1000$, $max\_depth = 8$
              \item $learning\_rate = 0.1$
              \item $feature\_fraction = 0.8$, $bagging\_fraction = 0.8$
              \item $random\_state = 42$
          \end{itemize}

    \item \textbf{Random Forest}: \texttt{RandomForestRegressor}
          \begin{itemize}
              \item $n_{\text{estimators}} = 500$
              \item $max\_depth = 15$
              \item $min\_samples\_split = 5$, $min\_samples\_leaf = 2$
              \item $n\_jobs = -1$, $random\_state = 42$
          \end{itemize}
\end{itemize}

\paragraph{Neural Network Regression Model}

\begin{itemize}
    \item architecture: hidden layers of sizes $512, 256, 128, 64$, each followed by ReLU, BatchNorm1d, Dropout($p=0.2$)
    \item output: Linear($64, 1$) + Sigmoid
    \item optimizer: Adam ($lr = 1\times 10^{-3}$, $weight\_decay = 1\times 10^{-5}$)
    \item batch size: 64
    \item loss: L1 Loss (MAE)
    \item early stopping: patience $= 15$, maximum $100$ epochs
\end{itemize}

%% file: Tables/DHS_variables.tex
\begin{table}[H]
    \caption{Summary of Variables and Definitions of Severe Deprivation}
    \centering
    \renewcommand{\arraystretch}{1.9} 
    \setlength{\tabcolsep}{5pt} 
    \begin{center}
        \begin{tabular}{|p{2.7cm}|p{2cm}|p{4.7cm}|p{7.3cm}|@{}m{0pt}@{}}
            \hline
            \textbf{Categories} & \textbf{Variables} & \textbf{Description} & \textbf{Definition of Severe Deprivation} \\
            \hline
            
            \multirow{2}{*}{Housing} 
            & hv216 & Sleeping rooms in household & \multirow{2}{7.3cm}{The sleeping density (hv216 / no. household) exceeds 5 persons per room.} \\
            \cline{2-3}
            & hv271 & Wealth index score &\\[0ex] 
            \hline
            
            \multirow{2}{*}{Water} 
            & hv201 & Main drinking water source & \multirow{2}{7.3cm}{The household's source of drinking water is from an unprotected well, spring, river, or other unprotected sources.} \\
            \cline{2-3}
            & hv204 & Time to water source &\\[0ex] 
            \hline
            
            \multirow{2}{*}{Sanitation}
                & hv205 & Type of toilet facility & \multirow{2}{7.3cm}{The household uses \textit{unimproved toilet facilities} or has \textit{no toilet facilities} at all.} \\
                \cline{2-3}
                & hv225 & Toilet sharing status &\\[0ex] 
            \hline
            
            Nutrition & hc70 & Height-for-age z-score &  The child’s height-for-age z-score is more than 3 standard deviations below.& \\[0ex] 
            \hline
            
            \multirow{7}{*}{Health}
                & h10 & Child received any vaccination 
                & \begin{center}
                    \multirow{7}{7.3cm}{Meeting any one of the following:
                        \begin{enumerate}
                            \item Children aged 12--35 months did not receive immunization against measles nor any dose of DPT.
                            \item Girls aged 15--17 years have an unmet need for contraception and are not using any contraceptive method.
                            \item Children experienced an acute respiratory infection (fever or cough) and received no treatment.
                        \end{enumerate}
                    }
                \end{center} \\
            
                \cline{2-3}
                & h3 & DPT 1 vaccination & \\
                \cline{2-3}
                & h5 & DPT 2 vaccination & \\
                \cline{2-3}
                & h7 & DPT 3 vaccination & \\
                \cline{2-3}
                & h9 & Measles 1 vaccination & \\
                \cline{2-3}
                & h31 & Child had cough recently & \\
                \cline{2-3}
                & v312 & Current contraceptive method &\\[0ex] 
            \hline
            
            \multirow{3}{*}{Education}
                & hv106 & Highest education level in household 
                & \makecell{\multirow{3}{7.3cm}{\RaggedRight Meeting any one of the following:
                        \begin{enumerate}
                            \item Children aged 6--14 who have never been to school.
                            \item Children aged 15--17 who have not completed primary school.
                        \end{enumerate}}}\\
                \cline{2-3}
                & hv109 & Educational attainment recoded & \\
                \cline{2-3}
                & hv121 & School attendance current year & \\
            \hline
            
            \multirow{2}{*}{New Indicators}
                & hv025 & Rural/Urban Status 
                & \multirow{2}{7.3cm}{New indicators.} \\
                \cline{2-3}
                & hv270 & Wealth index &\\[0ex] 
                
            \hline
        \end{tabular}
    \end{center}
\end{table}

%% file: Tables/kidsat_country.tex
\begin{table}[H]
    \caption{KidSat Country Summary}
    \centering
    \scriptsize
    \begin{tabularx}{\textwidth}{l l c l r r r l}
        \hline
        \textbf{KidSat Status} & \textbf{Country} & \textbf{Code} & \textbf{Region} & 
        \textbf{No\_Clusters} & \textbf{No\_Images} & \textbf{No\_Survey} & \textbf{Survey Year} \\
        \hline
        Original & Burundi & BU & Eastern Africa & 928 & 2784 & 2 & 2000, 2016 \\
        Original & Ethiopia & ET & Eastern Africa & 2561 & 7683 & 5 & 2000, 2005, 2010, 2016, 2019 \\
        Original & Kenya & KE & Eastern Africa & 4072 & 12216 & 4 & 2003, 2008, 2014, 2022 \\
        Original & Comoros & KM & Eastern Africa & 242 & 726 & 1 & 2012 \\
        Original & Madagascar & MD & Eastern Africa & 1235 & 3705 & 3 & 1997, 2008, 2021 \\
        Original & Malawi & MW & Eastern Africa & 2757 & 8271 & 4 & 2000, 2004, 2010, 2015 \\
        Original & Mozambique & MZ & Eastern Africa & 610 & 1830 & 1 & 2011 \\
        Original & Rwanda & RW & Eastern Africa & 2186 & 6558 & 5 & 2005, 2008, 2010, 2014, 2019 \\
        Original & Tanzania & TZ & Eastern Africa & 1694 & 5082 & 4 & 1999, 2010, 2015, 2022 \\
        Original & Uganda & UG & Eastern Africa & 1688 & 5064 & 4 & 2000, 2006, 2011, 2016 \\
        Original & Zambia & ZM & Eastern Africa & 1574 & 4722 & 3 & 2007, 2013, 2018 \\
        Original & Zimbabwe & ZW & Eastern Africa & 1189 & 3567 & 4 & 1999, 2005, 2010, 2015 \\
        Original & Angola & AO & Middle Africa & 625 & 1875 & 1 & 2015 \\
        Original & Lesotho & LS & Southern Africa & 1175 & 3525 & 3 & 2004, 2009, 2014 \\
        Original & Eswatini & SZ & Southern Africa & 270 & 810 & 1 & 2006 \\
        Original & South Africa & ZA & Southern Africa & 746 & 2238 & 1 & 2017 \\
        New & Congo Democratic Republic & CD & Middle Africa & 492 & 1476 & 1 & 2013 \\
        New & Cameroon & CM & Middle Africa & 1471 & 4413 & 2 & 2011, 2018 \\
        New & Gabon & GA & Middle Africa & 722 & 2166 & 2 & 2012, 2019 \\
        New & Chad & TD & Middle Africa & 624 & 1872 & 1 & 2014 \\
        New & Namibia & NM & Southern Africa & 1301 & 3903 & 1 & 2013 \\
        New & Burkina Faso & BF & Western Africa & 1452 & 4356 & 2 & 2010, 2021 \\
        New & Benin & BJ & Western Africa & 1290 & 3870 & 2 & 2012, 2017 \\
        New & Cote d'Ivoire & CI & Western Africa & 880 & 2640 & 2 & 2012, 2021 \\
        New & Ghana & GH & Western Africa & 1445 & 4335 & 2 & 2014, 2022 \\
        New & Guinea & GN & Western Africa & 992 & 2976 & 2 & 2012, 2018 \\
        New & Liberia & LB & Western Africa & 934 & 2802 & 2 & 2013, 2019 \\
        New & Mali & ML & Western Africa & 1562 & 4686 & 3 & 2016, 2012, 2018 \\
        New & Nigeria & NG & Western Africa & 3517 & 10551 & 2 & 2013, 2018 \\
        New & Niger & NI & Western Africa & 476 & 1428 & 1 & 2012 \\
        New & Sierra Leone & SL & Western Africa & 1342 & 4026 & 2 & 2008 \\
        New & Senegal & SN & Western Africa & 1441 & 4323 & 5 & 2016, 2017, 2018, 2019, 2023 \\
        New & Togo & TG & Western Africa & 330 & 990 & 1 & 2013 \\
        \hline
        \multicolumn{4}{l}{\textbf{Total}} & 43823 & 131469 & 79 & \\
        \hline
    \end{tabularx}
\end{table}